\documentclass[9pt,twocolumn,twoside]{nopnas-new}
\usepackage{comment}
\usepackage{fancyhdr}

\templatetype{pnasresearcharticle} 

\def\mic3proposal{IUS} 

\newcounter{notecounter}

\newcommand{\enoteson}{\long\gdef\enote##1##2{{
\stepcounter{notecounter}
{\large\bf
\hspace{1cm}\arabic{notecounter} $<<<$ ##1: ##2
$>>>$\hspace{1cm}}}}}
\enoteson

\renewenvironment{quote}{%
   \list{}{%
     \leftmargin0.5cm   
     \rightmargin\leftmargin
   }
   \item\relax
}
{\endlist}

\title{Extending Machine Language Models toward Human-Level Language Understanding}

\author[a,b,2]{James L. McClelland}
\author[b,2]{Felix Hill} 
\author[c,2]{Maja Rudolph}
\author[d,1,2]{Jason Baldridge}
\author[e,1,2]{Hinrich Sch\"{u}tze}

\affil[a]{Stanford University, Stanford, CA 94305, USA}
\affil[b]{DeepMind, London N1C 4AG, UK}
\affil[c]{Bosch Center for Artificial Intelligence, Renningen, 71272, Germany}
\affil[d]{Google Research, Austin, TX 78701, USA}
\affil[e]{LMU Munich, Munich, 80538, Germany}
\leadauthor{McClelland} 

\authorcontributions{JM, FH, MR, JB, and HS wrote the paper.}
\authordeclaration{The authors declare no conflict of interest.}
\equalauthors{\textsuperscript{1}J.B. and H.S. contributed equally.}
\correspondingauthor{\textsuperscript{2}
To whom correspondence should be addressed. E-mail: jlmcc\@stanford.edu, felixhill@google.com, marirudolph@gmail.com, 
jasonbaldridge@google.com or inquiries@cislmu.org}


\begin{abstract} 

Language is crucial for human intelligence, but what exactly is its role? We take language to be a part of a system for understanding and communicating about situations.
The human ability to understand and communicate about situations emerges gradually from experience and depends on domain-general principles of biological neural networks: connection-based learning, distributed representation, and context-sensitive, mutual constraint satisfaction-based processing. Current artificial language processing systems rely on the same domain general principles, embodied in artificial neural networks. Indeed, recent progress in this field depends on \emph{query-based attention}, which extends the ability of these systems to exploit context and has contributed to remarkable breakthroughs. Nevertheless, most current models focus exclusively on language-internal tasks, limiting their ability to perform tasks that depend on understanding situations. These systems also lack memory for the contents of prior situations outside of a fixed contextual span.  We describe the organization of the brain's distributed understanding system, which includes a fast learning system that addresses the memory problem. We sketch a framework for future models of understanding drawing equally on cognitive neuroscience and artificial intelligence and exploiting query-based attention. We highlight relevant current directions and consider further developments needed to fully capture human-level language understanding in a computational system.

\end{abstract}

\dates{This manuscript was compiled on \today}
\doi{\url{www.pnas.org/cgi/doi/10.1073/pnas.XXXXXXXXXX}}

\def\figref#1{Fig.~\ref{fig:#1}}
\def\figlabel#1{\label{fig:#1}\label{p:#1}}

\def\seclabel#1{\label{sec:#1}\label{p:#1}}
\def\eqref#1{Eq.~\ref{eqn:#1}}

\begin{document}
\thispagestyle{fancy}
\fancyhf{}
\fancyfoot[R]{\thepage} 
\fancyfoot[L]{McClelland et al.:
Extending Machine Language Models toward Human-Level
Language Understanding}

\maketitle
\ifthenelse{\boolean{shortarticle}}{\ifthenelse{\boolean{singlecolumn}}{\abscontentformatted}{\abscontent}}{}





Striking recent advances in machine intelligence have appeared in language tasks. Machines better transcribe speech and respond in ever more natural sounding voices. Widely available applications allow one to say something in one language and hear its translation in another. Humans perform better than machines in most language tasks, but these systems work well enough to be used by billions of people everyday. 

What underlies these successes? What limitations do they face?  We argue that progress has come from exploiting principles of neural computation employed by the human brain, while a key limitation is that these systems treat language as if it can stand alone.  We propose that language works in concert with other inputs to understand and communicate about situations.  We describe key aspects of human understanding and key components of the brain's understanding system. We then propose initial steps toward a model informed both by cognitive neuroscience and artificial intelligence and point to extensions addressing more abstract cases. 

\section*{Principles of Neural Computation}

The principles of neural computation are domain general, inspired by the human brain and human abilities. They were first articulated in the 1950s \cite{rosenblatt1961principles} and further developed in the 1980s in the Parallel Distributed Processing (PDP) framework for modeling cognition \cite{rm86}. This work introduced the idea that structure in cognition and language is \textit{emergent}: it is captured in learned connection weights supporting the construction of context-sensitive representations whose characteristics reflect a gradual, input-statistics dependent, learning process \cite{rm86past}. Classical linguistic theory and most computational linguistics employs discrete symbols and explicit rules to characterize language structure and relationships. In neural networks, these symbols are replaced by continuous, multivariate patterns called \textit{distributed representations} or \textit{embeddings} and the rules are replace by continuous, multi-valued arrays of connection weights that map patterns to other patterns. 

Since its introduction \cite{rm86past}, debate has raged about this approach to language processing \cite{pinker1988connections}.  Protagonists argue it supports nuanced, context- and similarity-sensitive processing that is reflected in the quasi-regular relationships between phrases and their sounds, spellings, and meanings \cite{macwhinney1991implementations,bybee2005alternatives}. These models also capture subtle aspects of human performance in language tasks \cite{seidenberg2014quasiregularity}. However, critics note that neural networks often fail to generalize beyond their training data, blaming these failures on the absence of explicit rules \cite{fodor1988connectionism,marcus2001algebraic,lake2017building}.

Another key principle is \textit{mutual constraint satisfaction} \cite{rumelhart77interactive}.  
For example, interpreting a sentence requires resolving both syntactic and semantic ambiguity. If we hear \textit{A boy hit a man with a bat}, we tend to assume \textit{with a bat} attaches to the verb (syntax) and thereby the instrument of hitting (semantics). However, if \textit{beard} replaces \textit{bat}, then \textit{with a beard} is attached to \textit{man} (syntax) and describes the person affected (semantics) \cite{taraban1988constituent}.  Even segmenting language into elementary units depends on meaning and context (\figref{rumelhart}). Rumelhart \cite{rumelhart77interactive} envisioned a model in which estimates of the probability of all aspects of an input constrain estimates of the probabilities of all others, motivating a model of context effects in perception \cite{mcclelland1981interactive} that launched the PDP approach. 



\section*{Neural Language Modeling}

\subsection*{Initial steps}  Elman \citep{elm90} introduced a simple recurrent neural network (RNN) (\figref{elman}a) that captured key characteristics of language structure through learning, a feat once considered impossible \cite{gold1967language}. It was trained to predict the next word in a sequence ($w(t+1)$) based on the current word ($w(t)$) and its own \textit{hidden} (that is, learned internal) representation from the previous time step ($h(t-1)$). Each of these inputs is multiplied by a matrix of connection weights (arrows labeled $W_{hi}$ and $W_{hh}$ in \figref{elman}a) and the results are added to produce the input to the hidden units.  The elements of this vector pass through a function limiting the range of their values, producing the hidden representation. This in turn is multiplied with weights to the output layer from the hidden layer ($W_{oh}$) to generate a vector used to predict the probability of each of the possible successor words. Learning is based on the discrepancy between the network's output and the actual next word; the values of the connection weights are adjusted by a small amount to reduce the discrepancy.  The network is \textit{recurrent} because the same connection weights (denoted by arrows in the figure) are used to process each successive word.

\begin{figure}
\begin{center}
\includegraphics[width=6.5cm]{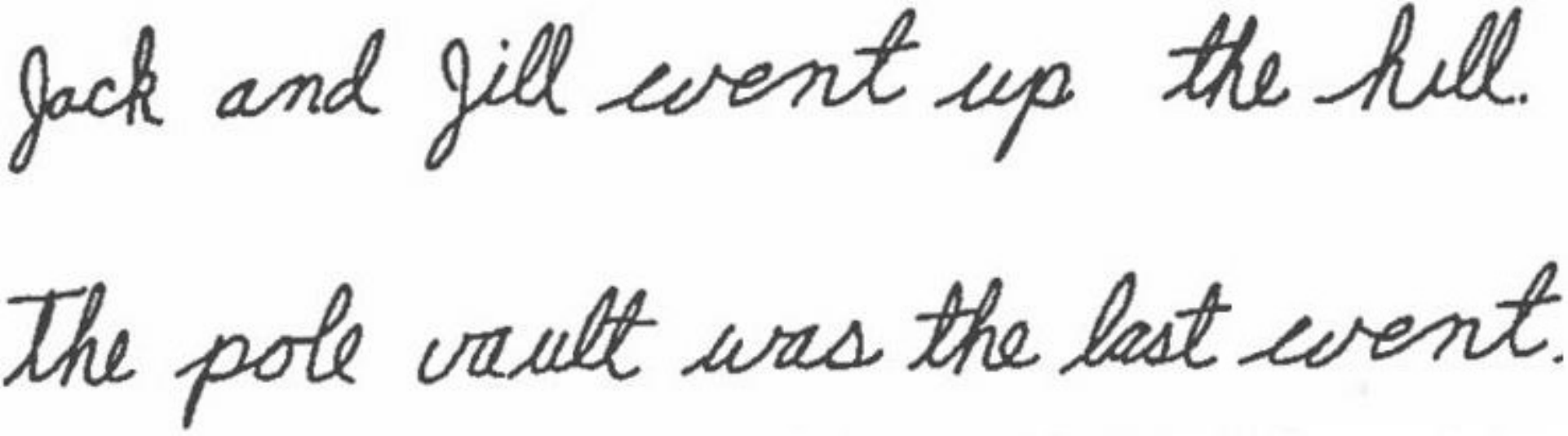}
\end{center}
\caption{\figlabel{rumelhart} Context influences the identification of letters in written text: the visual input we read as \textit{went} in the first sentence and \textit{event} in the second is the same bit of Rumelhart's handwriting, cut and pasted into each context.  Reprinted from \cite{rumelhart77interactive}.}
\end{figure}

Elman showed two things. First, after training his network to predict the next word in sentences like \textit{man eats bread}, \textit{dog chases cat}, and \textit{girl sleeps}, the network's representations captured the syntactic distinction between nouns and verbs \cite{elm90}. They also captured interpretable subcategories, as shown by a hierarchical clustering of the hidden representations of the different words (\figref{elman}b). This illustrates a key feature of learned representations: they capture specific as well as general or abstract information.  By using a different learned representation for each word, its specific predictive consequences can be exploited.  Because representations for words that make similar predictions are similar, and because neural networks exploit similarity, the network can share knowledge about predictions among related words. 

Second, Elman \cite{elm91} used both simple sentences like \textit{boy chases dogs} and more complex ones like \textit{boy who sees girls chases dogs}. In the latter, the verb \textit{chases} must agree with the first noun (\textit{boy}), not the closest noun (\textit{girls}), since the sentence contains a main clause (\textit{boy chases dogs}) interrupted by a reduced relative clause (\textit{boy [who] sees girls}). The model learned to predict the verb form correctly despite the intervening clause, showing that it acquired sensitivity to the syntactic structure of language, not just local co-occurrence statistics.

\subsection*{Scaling up to natural text}

Elman's task of predicting words based on context has been central to neural language modeling. However, Elman trained his networks with tiny, toy languages. For many years, it seemed they would not scale up, and language modeling was dominated by simple $n$-gram models and systems designed to assign explicit structural descriptions to sentences, aided by advances in probabilistic computations \cite{manning1999foundations}.  Over the past 10 years, breakthroughs have allowed networks to predict and fill in words in huge natural language corpora.

One challenge is the large size of a natural language's vocabulary. A key step was the introduction of methods for learning word representations (now called embeddings) from co-occurrence relationships in large text corpora \citep{collobert2011natural,mikolov2013distributed}. These embeddings exploit both general and specific predictive relationships of all the words in the corpus, improving generalization: task-focused neural models trained on small data sets better generalize to infrequent words (e.g., \textit{settee}) based on frequent words (e.g. \textit{couch}) with similar embeddings.

\begin{figure}[t]
\includegraphics[width=8.5cm]{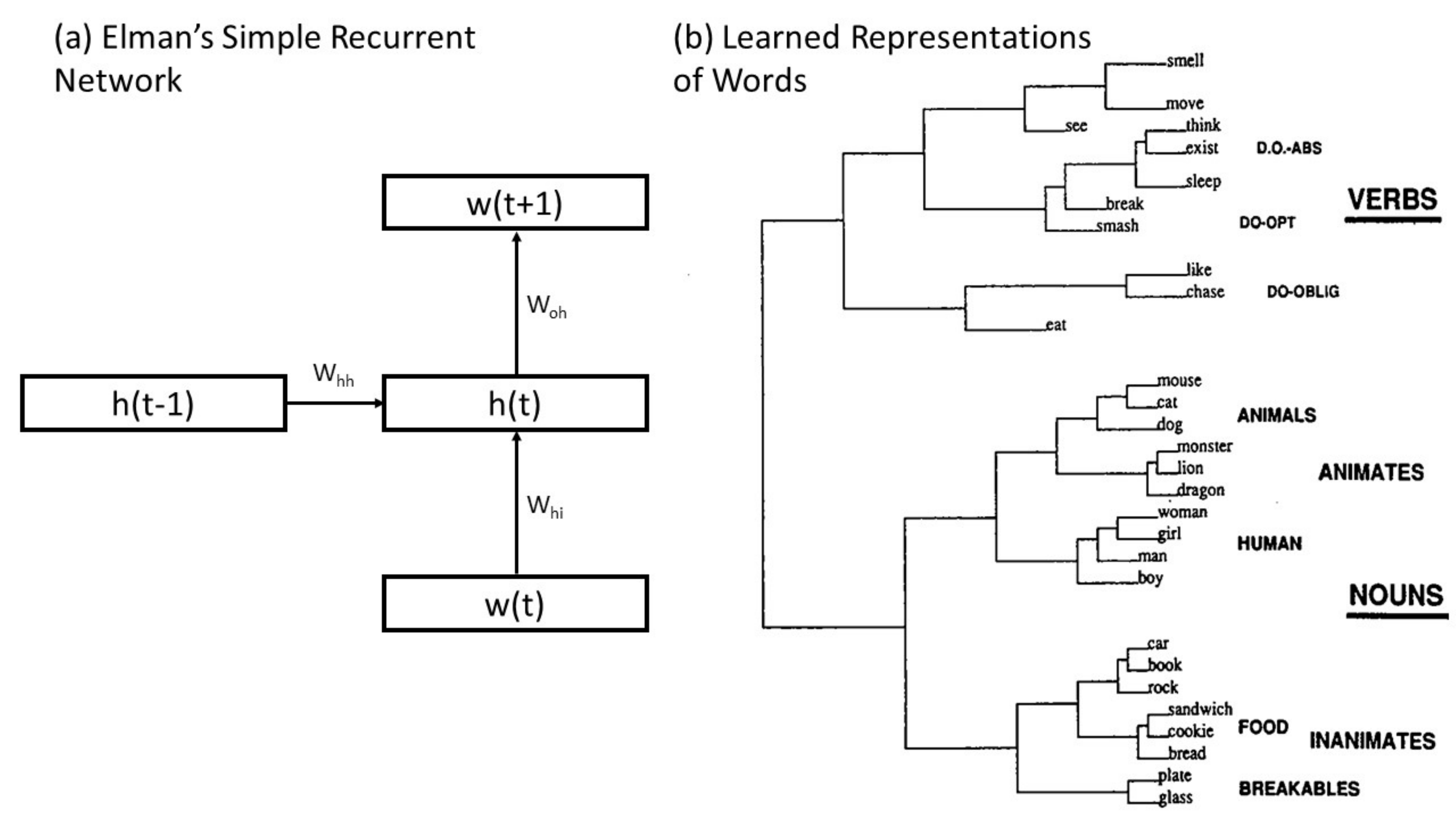}
\caption{\figlabel{elman} (a) Elman's (1990) simple recurrent network and (b) his hierarchical clustering of the representations it learned, reprinted from \cite{elm90}.}
\end{figure}

A second challenge is the indefinite length of the context that might be relevant for prediction. Consider this passage:

\begin{quote}
John put some beer in a cooler and went out with his friends to play volleyball. Soon after he left, someone \textit{took the beer} out of the cooler. John and his friends were thirsty after the game, and went back to his place for some beers.  When John opened the cooler, he discovered that the beer was \_\_\_.
\end{quote}
\noindent
Here a reader expects the missing word to be \textit{gone}. Yet if we replace \textit{took the beer} with \textit{took the ice}, the expected word is \textit{warm}.  Any amount of additional text between \textit{beer} and \textit{gone} does not change the predictive relationship, challenging RNNs like Elman's.  An innovation called \textit{Long-short-term memory} (LSTM) \cite{hochreiter1997long} partially addressed this problem by augmenting the recurrent network architecture with learned connection weights that gate information into and out of a network's internal state. However, LSTMs did not fully alleviate the context bottleneck problem \cite{Bahdanau2014}: a network's internal state was still a fixed-length vector, limiting its ability to capture contextual information.

\subsection*{Query-based attention}

Recent breakthroughs depend on an innovation we call \emph{query-based attention} (QBA) \cite{Bahdanau2014}. It was used in the Google Neural Machine Translation system \cite{wu16gnmt}, a system that attained a sudden leap in performance and attracted widespread public interest \cite{lewiskraus2016greatawakening}. 

\begin{figure}[t]
\begin{center}
\includegraphics[width=8.5cm]{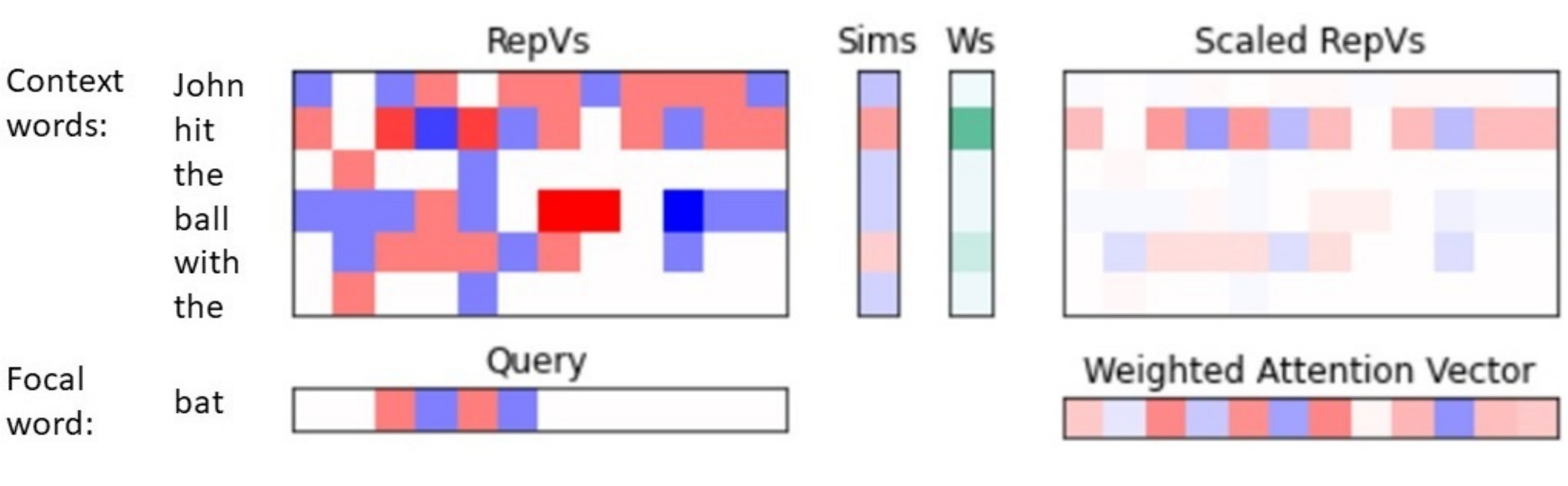}
\end{center}
\caption{\figlabel{atten-fig} Query-based attention (QBA). To constrain the interpretation of the word \textit{bat} in the context \textit{John hit the ball with the \_\_}, a \emph{query} generated from \textit{bat} is used to construct a \emph{weighted attention vector} which shapes the word's interpretation. The query is compared to each of the learned representation vectors (RepVs) of the context words; this creates a set of similarity scores (Sims) which in turn produce a set of weightings (Ws, a set of positive numbers summing to 1). The Ws are used to scale the RepVs of the context words, creating Scaled RepVs. The weighted attention vector is the element-wise sum of the Scaled RepVs. The Query, RepVs, Sims, Scaled RepVs and weighted attention vector use red color intensity for positive magnitudes and blue for negative magnitudes.  Ws are shown as green color intensity. White = 0 throughout.
The Query and RepVs were made up for illustration, inspired by \cite{manning2020emergent}. Mathematical details: For query $\mathbf{q}$ and representation vector $\mathbf{v_j}$ for context word $j$, the similarity score $s_j$ is $\cos{(q,v_{j})}$.  The $s_j$ are converted into weightings $w_j$ by the softmax function, $w_j{=}e^{(g s_j)} / (\large{\Sigma_{j\prime}} e^{(g s_{j\prime})})$, where the sum in the denominator runs over all words in the context span, and $g$ is a scale factor.}
\end{figure}

We illustrate QBA in \figref{atten-fig} with the sentence \textit{John hit the ball with the bat}.  Context is required to determine whether \textit{bat} refers to an animal or a baseball bat.  QBA addresses this by issuing \emph{queries} for relevant information. A query might ask ‘is there an action and relation in the context that would indicate which kind of bat fits best?’  The embeddings of words that match the query then receive high weightings in the weighted attention vector.  In our example, the query matches the embedding of \textit{hit} closely and of \textit{with} to some extent; the returned attention vector captures the content needed to determine that a baseball bat fits in this context.

There are many variants of QBA; the figure presents a simple version.
One important QBA model called BERT \cite{devlin} uses both preceding and following context. In BERT, the network is trained to correct missing or randomly replaced words in blocks of text, typically spanning two sentences. BERT relies on multiple attention heads, each employing QBA \cite{Vaswani2017}, concatenating the returned weighted vectors to form a composite vector.  The process is iterated across several stages, so that contextually constrained representations of each word computed at intermediate stages in turn constrain the representations of every other word in later stages. In this way, BERT employs mutual constraint satisfaction, as Rumelhart \cite{rumelhart77interactive} envisioned.  The contextualized embeddings that result from QBA can capture gradations within the set of meanings a language maps to a given word, aiding translation and other down-stream tasks. For example, in English, we use \textit{ball} for many types of balls, whereas in French, some are \textit{balles} and others \textit{ballons}.  Subtly different embeddings for \textit{ball} in different English sentences aids selecting the correct French translation. 

In QBA architectures, the vectors all depend on learned connection weights, and analysis
of BERT's representations shows that they capture syntactic structure well \cite{manning2020emergent}.  Different attention heads capture different linguistic relationships, and the similarity relationships among the contextually-shaded word representations can be used to reconstruct a sentence's syntactic structural description. 
These representations capture this structure \textit{without} building it in, supporting the emergence principle. That said, careful analysis \cite{linzen2020syntactic} indicates the deep network's sensitivity to grammatical structure is still imperfect, and only partially understood.

Some attention-based models \cite{brown2020language} proceed sequentially, predicting each word using QBA over prior context, while BERT uses parallel processing or mutual QBA simultaneously on all the words in a pair of sentences.  Humans appear to exploit past context and a limited window of subsequent context \cite{Warren392}, suggesting a hybrid strategy. Some machine models adopt this approach \cite{xlnet}, and below we adopt a hybrid approach as well.

Attention-based models have produced remarkable improvements on a wide range of language tasks. The models can be pre-trained on massive text corpora, providing useful representations for subsequent fine tuning to perform other tasks. A recent model called GPT-3 achieves impressive gains on several benchmarks without requiring fine-tuning \cite{brown2020language}.  However, this model still falls short of human performance on tasks that depend on what the authors call "common sense physics" and on carefully crafted tests of their ability to determine if a sentence follows from a preceding text \cite{nie2019adversarial}.  Further, the text corpora these models rely on are far larger than a human learner could process in a lifetime.  Gains from further increases may be diminishing, and human learners appear to be far more data-efficient. The authors of GPT-3 note these limitations and express the view that further improvements may require more fundamental changes.

\section*{Language in an Integrated Understanding System}

Where should we look for further progress addressing the limitations of current language models?  In concert with others \cite{bisk2020experience}, we argue that part of the solution will come from treating language as part of a larger system for understanding and communicating.

\subsubsection*{Situations}  We adopt the perspective that the targets of understanding are \textit{situations}. Situations are collections of entities, their properties and relations, and patterns of change in them. A situation can be static (e.g., a cat on a mat).  Situations include events (e.g., a boy hitting a ball). Situations can embed within each other; the cat may be on a mat inside a house on a particular street in a particular town, and the same applies to events like baseball games. A situation can be conceptual, social or legal, such as one where a court invalidates a law. A situation may even be imaginary. The entities participating in a situation or event may be real or fictitious physical objects or locations; animals, persons, groups or organizations; beliefs or other states of mind; sets of objects (e.g., all dogs); symbolic objects such as symbols, tokens or words; or even contracts, laws, or theories.  Situations can even involve changes in beliefs about relationships among classes of objects (e.g. biologists' beliefs about the genus a species of trees belongs in).  


What it means for an agent to understand a situation is to construct a representation of it that captures aspects of the participating objects, their properties, relationships and interactions, and resulting outcomes.
We emphasize that the understanding should be thought of as a \textit{construal} or interpretation that may be incomplete or inaccurate. The construal will depend on the culture and context of the agent and the agent's purpose. When other agents are the source of the input, the target agent's construal of the knowledge and purpose of these other agents also play important roles.  As such, the construal process must be considered to be completely open ended and to potentially involve interaction between the construer and the situation, including exploration of the world and discourse between the agent and participating interlocutors.

Within this construal of understanding, we emphasize that language should be seen as a component of an understanding system.  This idea is not new, but historically it was not universally accepted.  Chomsky, Fodor and others \cite{chomsky1971deep,fodor1983modularity,fodor1988connectionism} argued that grammatical knowledge sits in a distinct, encapsulated subsystem. Our proposal to focus on language as part of a system representing situations builds on a long tradition in linguistics \cite{lak87}, human cognitive psychology \cite{bransford1972contextual}, psycholinguistics \cite{crain1985context}, philosophy \cite{Montague:1973} and artificial intelligence \cite{schank1983dynamic}. The approach was adopted in an early neural network model \cite{STJOHN1990217} and aligns with other current perspectives in cognitive neuroscience \cite{hasson2018grounding} and artificial intelligence \cite{bisk2020experience}.

\subsubsection*{People construct situation representations}

A person processing language constructs a representation of the described situation in real time, using both the stream of words and other available information.  Words and their sequencing serve as \textit{clues to meaning} \cite{rumelhart1979problems} that jointly constrain the understanding of the situation \cite{STJOHN1990217}. 
Consider this passage:
\begin{quote}
    John spread jam on some bread.  The knife had been dipped in poison.
\end{quote}{}
We make many inferences: the jam was spread with the poisoned knife and poison has been transferred to the bread. If John eats it he may die! Note the entities are objects, not words, and the situation could be conveyed by a silent movie.

Evidence that humans construct situation representations from language comes from classic work by Bransford and colleagues \cite{bransford1972contextual,BARCLAY1974471}.  This work demonstrates that (1) we understand and remember texts better when we can relate the text to a familiar situation; (2) relevant information can come from a picture accompanying the text; (3) what we remember from a text depends on the framing context; (4) we represent objects in memory that were not explicitly mentioned; and (5) after hearing a sentence describing spatial or conceptual relationships, we remember these relationships, not the language itself. For example, given \textit{Two frogs rested beside a floating log and a fish swam under it}, the situation changes if \textit{it} is replaced by \textit{them}. After hearing the original sentence, people reject the variant with \textit{it} in it as the sentence they heard before, but if the initial sentence said the frogs rested \textit{on} the log, the situation is unchanged by replacing \textit{it} with \textit{them}, and people accept this variant.

Evidence from eye movements shows that people use linguistic and non-linguistic input jointly and immediately \cite{tanenhaus1995integration}. Just after hearing \textit{The man will drink ...} participants look at a full wine glass rather than an empty beer glass \cite{altmann2007real}. After hearing \textit{The man drank}, they look at the empty beer glass. Understanding thus involves constructing, in real time, a representation conveyed jointly by vision and language.


\subsection*{The compositionality of situations}

An important debate in cognitive science and AI centers on compositionality.  Fodor and Pylyshyn \cite{fodor1988connectionism} argued that our cognitive systems must be compositional by design to allow language to express arbitrary relationships and noted that early neural network models failed tests of compositionality. Such failures are still reported \cite{lake2018generalization} leading some to propose building compositionality in \cite{lake2017building}; yet, as we have seen, the most successful language models avoid doing so.  We suggest that a focus on situations may enhance compositionality because situations are themselves compositional.  Suppose a person picks an apple and gives it to someone. A small number of objects and persons are focally involved, and the effects on other persons and objects are likely to be local.  A sentence like \textit{John picked an apple and gave it to Mary} could describe this situation, capturing the most relevant participants and their relationships.  We emphasize that compositionality is predominant and approximate, not universal or absolute, so it is best to allow for these matters of degree.  Letting situation representations emerge through experience will help our models to achieve greater systematicity, while leaving them with the flexibility that has led to their successes to date.

\subsection*{Language informs us about situations} 

Situations ground the representations we construct from language; equally importantly, language informs us about situations.  Language tells us about situations we have not witnessed and describes aspects that we cannot observe.  Language also communicates folk or scientific construals that shape listener's construals, such as the idea that an all-knowing being took six days to create the world or the idea that natural processes gave rise to our world and ultimately ourselves over billions of years.  Language can be used to communicate information about properties that only arise in a social setting, such as ownership, or that have been identified by a culture as important, such as exact number.  Language thus enriches and extends the information we have about situations and provides the primary medium conveying properties of many kinds of objects and many kinds of relationships.

\section*{Toward a Brain and AI Inspired Model of Understanding}

Capturing the full range of situations is clearly a long-term challenge. We return to this later, focusing first on concrete situations involving animate beings and physical objects. We seek to integrate insights from cognitive neuroscience and artificial intelligence toward the goal of building an integrated understanding model.  We start with our construal of the understanding system in the human brain and then sketch aspects of what an artificial implementation might look like.

\subsection*{The understanding system in the brain}

\seclabel{cls}

\begin{figure}[t]
\begin{center}
\includegraphics[width=7cm]{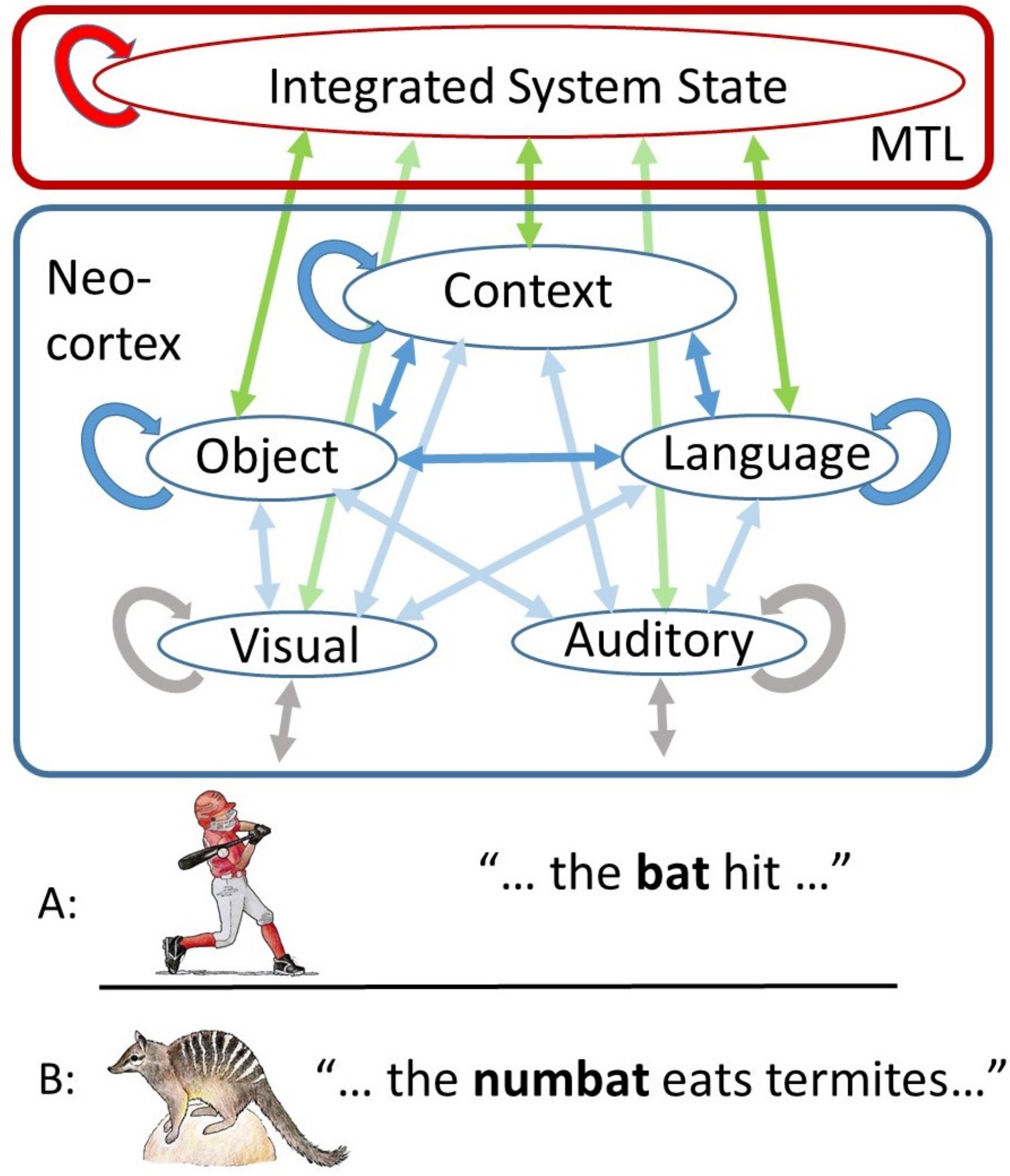}
\end{center}
\caption{\figlabel{jay-combined} Sketch of the brain's understanding system.  Ovals in the blue box stand for neocortical brain areas representing different kinds of information. Arrows in the neocortex stand for connections allowing representations to constrain each other.  The medial temporal lobe (red box) stores an integrated representation of the neocortical system state arising from an experience. The red arrow represents fast-learning connections that store this pattern for later reactivation and use.  Green arrows stand for gradually learned connections supporting bidirectional influence between the MTL and neocortex. (A) and (B) are two example inputs discussed in the main text.}
\end{figure}

Our construal of the human integrated understanding system builds on the principles of mutual constraint satisfaction and emergence and with the idea that understanding centers on the construction of situation representations.  It is consistent with a wide range of evidence, some of which we review, and is broadly consistent with recent characterizations in cognitive neuroscience \cite{Ranganath2012,hasson2018grounding}.  However, researchers hold diverse views about the details of these systems and how they work together.

We focus first on the part of the system located primarily in the neocortex of the brain, as schematized in the large blue box of \figref{jay-combined}. Together with input and output systems, this allows a person to combine linguistic and visual input to understand the situation referred to upon hearing a sentence, such as one containing the word \textit{bat}, while observing a corresponding situation in the world. It is important to note that the neocortex is very richly structured, with on the order of 100 well-defined anatomical subdivisions. However, it is common and useful to group these divisions into subsystems. The ones we focus on here are each indicated by a blue oval in the figure. One subsystem subserves the formation of a visual representation of the given situation, and another subserves the formation of an auditory representation capturing the spatiotemporal structure of the co-occurring spoken language. The three ovals above these provide representations of more integrative/abstract types of information (see below). 

Within each subsystem, and between each connected pair of subsystems, the neurons are reciprocally interconnected via learning-dependent pathways allowing mutual constraint satisfaction among all of the elements of each of the representation types, as indicated by the looping blue arrows from each oval to itself and by the bi-directional blue arrows between these ovals.  Brain regions for representing visual and auditory inputs are well-established, and the evidence for their involvement in a mutual constraint satisfaction process with more integrative brain areas has been reviewed elsewhere \cite{mcclelland2014interactive,heilbron2020word}. Here we consider the three more integrative subsystems.

\subsubsection*{Object representations}

A brain area near the front of the temporal lobe houses neurons whose activity provides an embedding capturing the properties of objects \cite{patterson2007you}. Damage to this area impairs the ability to name objects, to grasp them correctly for their intended use, to match them with their names or the sounds they make, and to pair objects that go together, either from their names or from pictures.  This brain area is itself an inter-modal area, receiving visual, language and other information about objects such as the sounds they make and how they feel to the touch. Models capturing these findings \cite{rogers2004structure} treat this area as the hidden layer of an interactive, recurrent network with bidirectional connections to other layers representing different types of object properties, including the object's name.  In these models, an input to any of these other layers activates the corresponding pattern in the hidden layer, which in turn activates the corresponding patterns in the other layers. This supports, for example, the ability to produce the name of an object from visual input. Damage (simulated by removing neurons in the hidden layer) degrades the model's representations, capturing the patterns of errors made by patients with the condition.

\subsubsection*{Representation of context}

There is a network of areas in the brain that capture the large-scale spatiotemporal context of fine-grained static situations and short time scale events (hereafter \textit{micro-situations}) that can be conveyed by a sentence like \textit{the boy hit the ball with a bat}. These context representations arise in a set of interconnected areas primarily within the parietal lobes \cite{Ranganath2012,hasson2018grounding}. 
In recent work, brain imaging data is used to analyze the time-varying patterns of neural activity while processing a temporally extended narrative. The brain activity patterns that represent scenes extending over tens of seconds (e.g., a detective searching a suspect's apartment for evidence) are largely the same, whether the information comes from watching a movie, hearing or reading a narrative description, or recalling the movie after having seen it \citep{zadbood17transmitmemories,Baldassano2017}.  Activations in different brain areas track information at different time scales.  Activity in modality-specific areas associated with speech and visual processing follows the moment-by-moment time course of spoken and/or visual information while activity in the network associated with situation representations fluctuates on a much longer time scale.  During processing of narrative information, activations in these regions tends to be relatively stable within a scene and punctuated with larger changes at boundaries between these scenes, and these patterns lose their coherence when the narrative structure is scrambled \cite{hasson2018grounding,Baldassano2017}.  Larger-scale spatial transitions (e.g., transitions between rooms) also create large changes in neural activity \cite{Ranganath2012}.

\subsubsection*{The role of language}  Where in the brain should we look for representations of the relations among the objects participating in a micro-situation?
The effects of brain damage suggest that representations of relations may be integrated, at least in part, with the representation of language itself.  Injuries affecting the lateral surface of the frontal and parietal lobes produce profound deficits in the production of fluent language, but can leave the ability to read and understand concrete nouns largely intact.  Such lesions produce intermediate degrees of impairment to abstract nouns, verbs, and modifiers, and profound impairment to words like \textit{if} and \textit{by} that capture grammatical and spatial relations \cite{morton1980little}.  This striking pattern is consistent with the view that language itself is intimately tied to the representation of relations and changes in relations (information conveyed by verbs, prepositions, and grammatical markers). Indeed, the frontal and parietal lobes are associated with representation of space and action (which causes change in relations), and patients with lesions to the frontal and parietal language-related areas have profound deficits in relational reasoning tasks \cite{baldo2010relational}.  We therefore tentatively suggest that the understanding of micro-situations depends jointly on the object and language systems, and that the language is intimately link to representation of spatial relationships and actions.

\subsubsection*{Complementary learning systems}
\seclabel{jay-learning}

The brain systems described above support understanding of situations that draw on general knowledge as well as oft-repeated personal knowledge, but they do not support the formation of new memories that can be accessed and used at an arbitrary later time. This ability depends on structures that include the hippocampus in the medial temporal lobes (MTL; \figref{jay-combined}, red box).  While these areas are critical for new learning, damage to them does not affect general knowledge, acquired skills, or the ability to process language and other forms of input to understand a situation---except when this depends on remote information experienced briefly outside the immediate current context \cite{MILNER1968215}.  These findings are captured in the neural-network based complementary learning systems (CLS) theory \cite{Marr71archicortex,mcclelland95cls,kumaran16cls}, which holds that connections within the neocortex acquire the knowledge that allows a human to understand objects and their properties, to link words and objects, and to understand and communicate about generic situations as these are conveyed through language and other forms of experience.  According to this theory, the learning process is necessarily gradual, allowing it to capture the nuanced statistical structure of experience. The MTL provides a complementary fast-learning system supporting the formation of new arbitrary associations, linking the elements of an experience together, including the objects and language encountered in a situation and the co-occurring spatiotemporal context, as might arise in the situation depicted in \figref{jay-combined}B, where a person encounters the word \textit{numbat} from both visual and language input.

It is generally accepted that knowledge that depends initially on the MTL can be integrated into the neocortex through a consolidation process \cite{MILNER1968215}. 
In CLS \cite{mcclelland95cls}, the neocortex learns arbitrary new information gradually through interleaved presentations of new and familiar items, weaving it into the fabric of knowledge and avoiding interference with existing knowledge \cite{mcclelland2020integration}. The details are subjects of current debate and ongoing investigation \cite{yonelinas2019}.

As in our example of the beer John left in the cooler, understanding often depends on remote information. People exploit such information during language processing \cite{Meneti09Elephants}, and patients with MTL damage have difficulty understanding or producing extended narratives \cite{zuo2020temporal}; they are also profoundly impaired in learning new words for later use \cite{gabrieli1988impaired}. Neural language models, including those using QBA, also lack these capabilities. In BERT and related models, bi-directional attention operates within a span of a couple of sentences at a time.  Other models \cite{brown2020language} employ QBA over longer spans of prior context, but there is still a finite limit.  These models learn gradually like the human neocortical system, allowing them to capture structure in experience and acquire knowledge of word meaning. GPT-3 \cite{brown2020language} is impressive in its ability to use a word encountered for the first time within its contextual span appropriately in a subsequent sentence, but this information is lost forever when the context is re-initialized, as it would be in a patient without the medial temporal lobe.  Including an MTL-like system in future understanding models would address this limitation.

The brain's complementary learning systems may provide a means to address the challenge of learning to use a word encountered in a single context appropriately across a wide range of contexts.  Deep neural networks that learn gradually through many repetitions of an item that occurs in a single context, interleaved with presentations of other items occurring in a diversity of contexts, do not show this ability \cite{lake2018generalization}. We attribute this failure to the fact that the distribution of training experiences they receive conveys the information that the target item is in fact restricted to its single context.  Further research should explore whether augmenting a model like GPT-3 with an MTL-like memory would enable more human-like extension of a novel word encountered just once to a wider range of contexts.

\subsection*{Next steps toward a brain and AI-inspired model}

Given the construal we have described of the human understanding system, we now ask, what might an implementation of a model consistent with it be like?  This is a long-term research question -- addressing it will benefit from a greater convergence of cognitive neuroscience and AI.  Toward this goal, we sketch a proposal for a brain and AI-inspired model.  We rely on the principles of mutual constraint satisfaction and emergence, the query-based attention architecture from AI and deep learning, and the components and their interconnections in the understanding system in the brain, as illustrated in \figref{jay-combined}.  

We treat the system as one that receives sequences of picture-description (PD) pairs grouped into episodes that in turn form story-length narratives, with the language conveyed by text rather than speech.  Our sketch leaves important issues open and will require substantial development. Steps toward addressing some of these issues are already being taken: mutual QBA is already being used, e.g., in \cite{SunetalCoRR-19b}, to exploit audio and visual information from movies. 

In our proposed model, each PD pair is processed by interacting object and language subsystems, receiving visual and text input at the same time.  Each subsystem must learn to restore missing or distorted elements (words in the language subsystem, objects in the object subsystem) by using mutual QBA as in BERT, allowing each element in each subsystem to be constrained by all of the elements in both sub-systems. Additionally these systems will query the context and memory subsystems, as described below.

The context subsystem encodes a sequence of compressed representations of the previous PD pairs within the current episode. Processing in this subsystem would be sequential over pairs, allowing the network constructing the current compressed representation to query the representations of past pairs within the episode. Within the processing of a pair, the context system would engage in mutual QBA with the object and language subsystems, allowing the language and object subsystems to indirectly exploit information from past pairs within the episode.

Our system also includes an MTL-like memory to allow it to use remote information beyond the current episode. A neural network with learned connection weights constructs a reduced description of the state of the object, language, and context modules along with their inputs from vision and text after processing each PD pair. Building on existing artificial systems with external memory \cite{weston2015memory,graves2016hybrid} this compressed vector is stored as a vector in a slot in an external memory.  These states are then accessible to the cortical subsystems via QBA.  The system would employ a flexible querying scheme, such that any subset of the object, language, or context representations of an input currently being processed would contribute to accessing relevant MTL representations.  There contents would then be to all of the cortical subsystems using QBA.  Thus, the appearance of a numbat in a visual scene would retrieve the corresponding language and context information containing its name and the fact that it eats termites, based on prior storage of the the compressed representation formed previously from the inputs in \figref{jay-combined}B.

Ultimately, our model may benefit from a subsystem that guides processing in all of the subsystems we have described.  The brain has such a system in its frontal lobes; damage to this system leads to impairments in guiding behavior and cognition according to current task demands, and others advocate including such a system in neural AI systems \cite{russindeep}.  We leave it to future work to consider how to integrate such a subsystem into the model we have described here.

\subsection*{Enhancing understanding by incorporating interaction with the physical and social world}
\seclabel{grounded}

A complete model of the human understanding system will require integration of many additional information sources.  These include sounds, touch and force-sensing, and information about one's own actions.  Every source provides opportunities to predict information of each type, relying on every other type. Information salient in one source can bootstrap learning and inference in the other, and all are likely to contribute to enhancing compositionality and addressing the data-inefficiency of learning from language alone.  This affords the human learner an important opportunity to experience the compositional structure of the environment through its own actions. Ultimately, an ability to link one's actions to their consequences as one behaves in the world should contribute to the emergence of, and appreciation for, the compositional structure of events, and provide a basis for acquiring notions of cause and effect, of agency, and of object permanence \cite{piaget1952origins}.

These considerations motivate recent work on agent-based language learning in simulated interactive 3D environments~\cite{hermann2017grounded,das2017question,chaplot2018gated,oh2017zero}. In~\citep{hill2019emergent}, an agent was trained to identify, lift, carry and place objects relative to other objects in a virtual room, as specified by simplified language instructions. At each time step, the agent received a first-person visual observation and processed the pixels to obtain a representation of the scene. This was concatenated to the final state of an LSTM that processed the instruction, and then passed to an integrative LSTM whose output was used to select a motor action. The agent gradually learned to follow instructions of the form~\emph{find a pencil}, ~\emph{lift up a basketball} and~\emph{put the teddy bear on the bed}, encompassing 50 objects and requiring up to $70$ action steps. Such instructions require constructing representations based on language stimuli that support identification of objects and relations across space and time and the integration of this information to inform motor behaviors. 

Importantly, without building in explicit object representations, the learned system was able to interpret novel instructions. For instance, an agent trained to lift each of 20 objects, but only trained to put 10 of those in a specific location could place the remaining objects in the same location on command with over 90\% accuracy, demonstrating a degree of compositionality in its behavior. Notably, the agent's ego-centric, multimodal and temporally-extended experience contributed to this outcome;  both an alternative agent with a fixed perspective on a 2D grid world and a static neural network classifier that received only individual still images exhibited significantly worse generalization.
This underscores how affording neural networks access to rich, multi-modal interactive environments can stimulate the development of capacities that are essential for language learning, and contribute toward emergent compositionality.


\subsection*{Beyond concrete situations} 

How might our approach be extended beyond concrete situations to those involving relationships among objects like laws, belief systems, and scientific theories?  Basic word embeddings themselves capture some abstract relations via vector similarity, e.g., encoding that \textit{justice} is closer to \textit{law} than \textit{peanut}. Words are uttered in real world contexts and there is a continuum between grounding and language-based linking for different words and different uses of words. For example, \textit{career} is not only linked to other abstract words like \textit{work} and \textit{specialization} but also to more concrete ones like \textit{path} and its extended metaphorical use as the means to achieve goals \citep{bryson:2008}. Embodied, simulation-based approaches to meaning \citep{Lakoff/Johnson:1980,feldman:narayanan:2004} build on this observation to bridge from concrete to abstract situations via metaphor. They posit that understanding words like \textit{grasp} is linked to neural representations of the action of grabbing and that this circuitry is recruited for understanding contexts such as \textit{grasping an idea}. We consider situated agents as a critical catalyst for learning about how to represent and compose concepts pertaining to spatial, physical and other perceptually immediate phenomena---thereby providing a grounded edifice that can connect both to brain circuitry for motor action and to representations derived primarily from language.

\section*{Conclusion}
Language does not stand alone. The understanding system in the brain connects language to representations of objects and situations and enhances language understanding by exploiting the full range of our multi-sensory experience of the world, our representations of our motor actions, and our memory of previous situations.  We believe next generation language understanding systems should emulate this system and we have sketched an approach that incorporates recent machine learning breakthroughs to build a jointly brain and AI inspired understanding system.  We emphasize understanding of concrete situations and argue that understanding abstract language should build upon this foundation, pointing toward the possibility of one day building artificial systems that understand abstract situations far beyond concrete, here-and-now situations.  In sum, combining insights from neuroscience and AI will take us closer to human-level language understanding.


\acknow{This article grew out of a
workshop organized by HS at \textit{Meaning in Context 3}, Stanford University, September 2017.  We thank Janice Chen, Chris Potts, and Mark Seidenberg for discussion. HS was supported by ERC Advanced Grant \#740516.}

\showacknow{} 
\bibliography{ExtendingMLModels}

\end{document}